\documentclass[10pt,twocolumn,letterpaper]{article}

\usepackage{iccv}
\usepackage{times}
\usepackage{epsfig}
\usepackage{graphicx}
\usepackage{amsmath}
\usepackage{amssymb}
\usepackage{mathtools}

\usepackage{mathptmx}
\usepackage[scaled=0.92]{helvet}
\usepackage{microtype}
\usepackage{verbatim}
\usepackage{multirow}

\usepackage{array}
\usepackage{booktabs} 
\usepackage{multirow}
\usepackage{verbatim}

\usepackage[linesnumbered,ruled,vlined]{algorithm2e}
\usepackage{lipsum}

\usepackage[breaklinks=true,bookmarks=false]{hyperref}

\iccvfinalcopy % *** Uncomment this line for the final submission

\ificcvfinal\fi

\begin{document}

\title{Activation Template Matching Loss for Explainable Face Recognition}

\author{Huawei Lin \qquad Haozhe Liu \qquad Qiufu Li \qquad Linlin Shen* \\
\\
\normalsize Computer Vision Institute, Shenzhen University, Shenzhen 518060, China\\
\normalsize Shenzhen Institute of Artificial Intelligence and Robotics for Society, Shenzhen 518060, China \\
\normalsize{Guangdong Key Laboratory of Intelligent Information Processing, Shenzhen University, Shenzhen 518060, China}\\
{\tt\small huaweilin.cs@gmail.com} \qquad {\tt\small liuhaozhe2019@email.szu.edu.cn}\\
{\tt\small \{liqiufu, llshen\}@szu.edu.cn} 
}

\maketitle
\ificcvfinal\thispagestyle{empty}\fi

\begin{abstract}
Can we construct an explainable face recognition network able to learn a facial part-based feature like eyes, nose, mouth and so forth, without any manual annotation or additionalsion datasets? In this paper, we propose a generic Explainable Channel Loss (ECLoss) to construct an explainable face recognition network. The explainable network trained with ECLoss can easily learn the facial part-based representation on the target convolutional layer, where an individual channel can detect a certain face part. Our experiments on dozens of datasets show that ECLoss achieves superior explainability metrics, and at the same time improves the performance of face verification without face alignment. In addition, our visualization results also illustrate the effectiveness of the proposed ECLoss. 
\end{abstract}

\section{Introduction}
Face recognition has made significant achievement in the past ten years \cite{parkhi2015deep, schroff2015facenet, deng2019arcface, kim2020groupface, jiang2021pointface, yang2021ramface}, due to the development of deep learning, large-scale labeled faces datasets and so forth, and has been extensively used in many applications \cite{huang2021one, xia2020analyzing}. Despite of the great success, most of these face recognition algorithms are still "black-box".

What face recognition networks have learned? It is commonly believed that the shallow layers are more likely to capture low-level information like colors, edges, and textures, while deep layers tend to describe high-level features \cite{zeiler2014visualizing, lin2017feature}. However, such observation is too general to understand a face recognition network. 

\begin{figure}[t]
	\includegraphics[width=\linewidth]{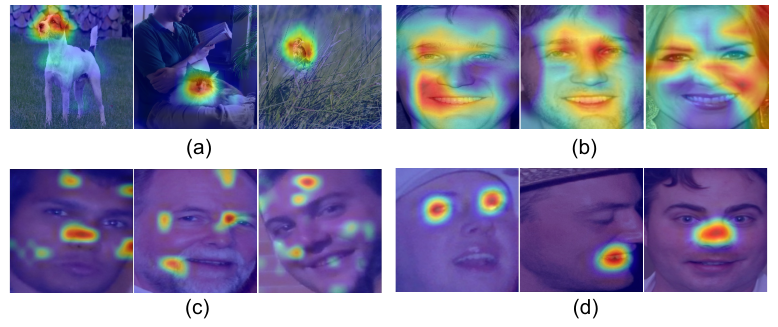}
	\centering
	\caption{Comparison between different methods.\textbf{(a)} In object classification, the visualization of Grad-CAM\cite{selvaraju2017grad} shows reasonable network prediction. The maximum activation regions (red) tend to cover the actual objects. \textbf{(b)} In face recognition,
		The maximum activation regions visualized by Grad-CAM\cite{selvaraju2017grad} focus on the whole face mostly, and does not provide effective evidence for explainability. \textbf{(c)} Feature map visualization for face recognition. The pattern of activation is random, and can not be understood by humans. \textbf{(d)} Feature map visualization for our explainable face recognition. A channel of feature maps represents a certain face part.}
	\label{fig:cam_cmp}
\end{figure}

A typical approach of previous studies for this problem is to use salience map to understand a network when it makes a decision \cite{bach2015pixel, zhou2016learning, fong2017interpretable, selvaraju2017grad, liu2021group}. However, most salience map based methods are not suitable for face recognition, as the areas of maximum activation are focused on the whole face as shown in Figure \ref{fig:cam_cmp}, which does not provide effective evidence and semantic information for explainability. Prior works have demonstrated that the representation extracted by deep networks may be mixed with high-dimensional abstract features that humans cannot understand. 

One potential solution for this issue is to guide the networks to learn a structural representation \cite{yin2019towards}. Yin {\em et al.} \cite{yin2019towards} suggested that in explainable face recognition network, each channel shall be able to represent a face part, and proposed a spatial activation diversity loss to learn an explainable representation. However, in order to learn explainable representation patterns, occluded faces are required for training. This additional step undoubtedly increases the difficulty of training. How can we find out a model-agnostic method to construct an explainable face recognition network without any additional occlusion face dataset?

\begin{figure*}[!thb]
	\includegraphics[width=\linewidth]{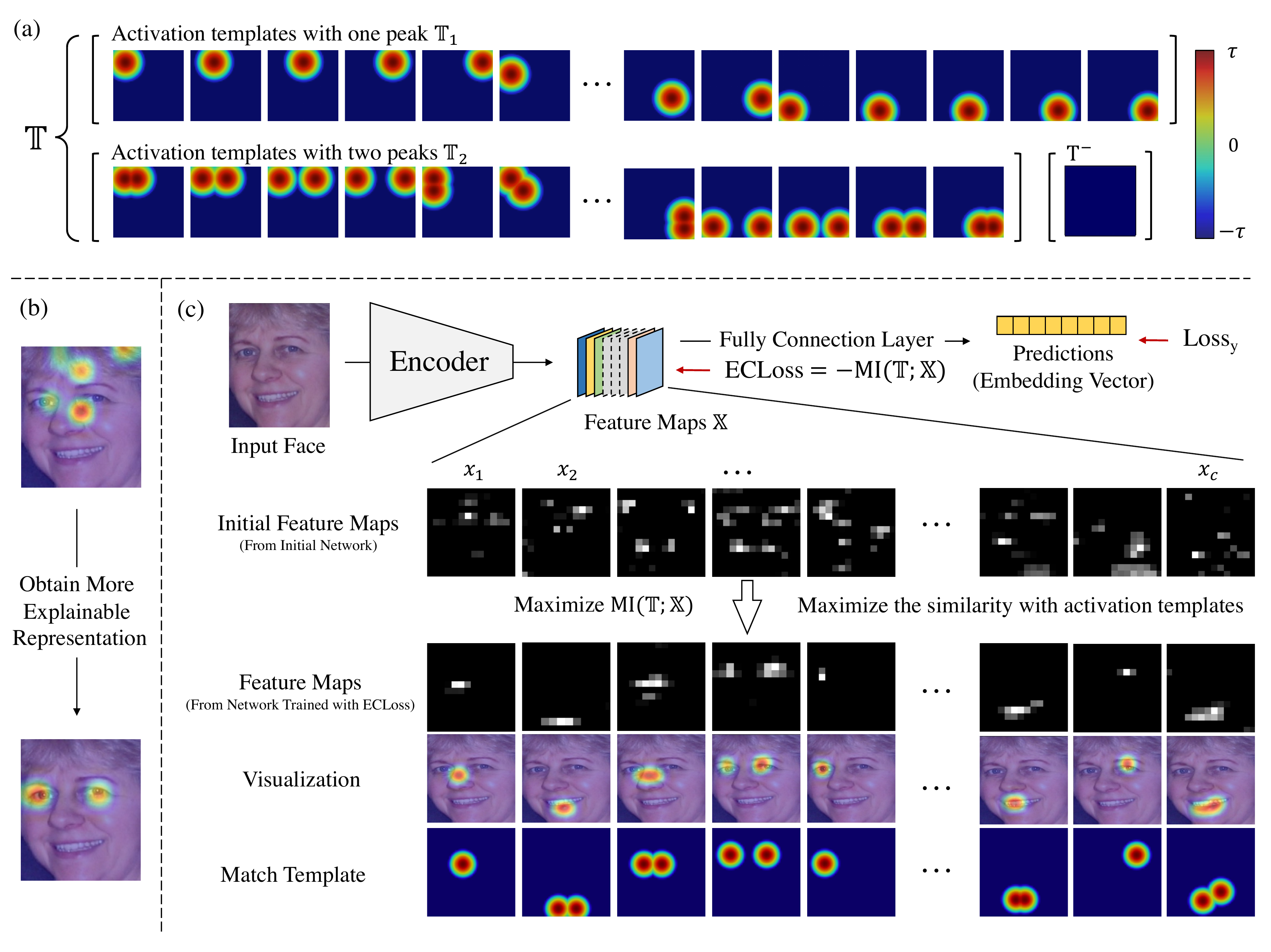}
	\centering
	\caption{The training pipeline of explainable network with the proposed ECLoss. {\bf (a)} The activation templates $\mathbb{T}$ are used for ECLoss to force the representations of the target convolutional layer to contain maximum two activation peaks. {\bf (b)} Compared with the network trained without ECLoss, the explainable network trained with ECLoss can obtain more explainable representation on the target convolution layer. {\bf (c)} The pipeline of training. The proposed ECLoss used mutual information, to force the target convolutional layer to represent the patterns of the activation templates.}
	\label{fig:channel_list}
\end{figure*}

Inspired by previous studies on the explainability and interpretability of deep learning \cite{zhang2018interpretable, yin2019towards, bau2017network, zhou2016learning, bau2017network, zhou2018interpreting, bau2020understanding, liu2022robust}, we proposed a generic Explainable Channel Loss (ECLoss) to construct an explainable face recognition network. Figure \ref{fig:cam_cmp} shows the difference in feature maps among different methods. By adopting ECLoss, the recognition network can stand a more readable representation for its decision behavior, which draw a remarkable progress over the existing explainable solutions. As shown in Figure \ref{fig:channel_list}, a channel of the explainable face recognition network trained with ECLoss can learn semantic parts like eyes, mouth, nose, {\em etc}. To the best of our knowledge, this is the first method to construct a feature level explainable face recognition network that does not require any additional dataset or manual annotation. In this way, the explainable network is able to effectively demonstrate what has been learned.

In conclusion, the proposed Explainable Channel Loss (ECLoss) has the following properties:

1) Networks trained  with ECLoss can explain what is learned inside the network. As shown in Figure \ref{fig:fig_list}, the visualization results show that different channels learned different semantic parts.

2) As shown in Figure \ref{fig:channel_list} and Figure \ref{fig:fig_list}, a single channel within the explainable face recognition network trained with ECLoss can learn different semantic parts like eyes, mouth, nose, {\em etc.}

3) Compared to \cite{zhang2018interpretable}, the training of ECLoss in the network is simple and effective, and does not require any manual annotation. {\em i.e.} labeling of nose, mouth, or other parts are not necessary.

4) As a model-agnostic method, ECLoss can be applied to various backbones of deep learning networks.

5) As an exploratory study, ECLoss achieved superior explainability metrics, and improved the performance of face verification.

\section{Related Work}
\subsection{Face Recognition}
Face recognition has been studied for several decades. Before deep learning, many previous studies are based on feature engineering \cite{cao2010face, kumar2009attribute, berg2013poof}. Due to the rapid development of deep learning, face recognition has made a huge achievement \cite{zhang2021face, sun2014deep, taigman2014deepface, parkhi2015deep, wang2017normface, schroff2015facenet, wen2016discriminative}. Sun {\em et al.} proposed a network to learn high-level feature representations through multi-class face identification tasks \cite{sun2014deep}. Taigman {\em et al.} proposed DeepFace, a nine-layer deep neural network, for large-scale face recognition \cite{taigman2014deepface}. Parkhi {\em et al.} constructed a deeper neural network named VGGFace and achieved high accuracy \cite{parkhi2015deep}. Wang {\em et al.} presented NormFace using normalized features for training, which achieved a higher accuracy than that without normalization \cite{wang2017normface}. Kim {\em et al.} presented GroupFace to improve the quality of the embedding vector by utilizing diverse group-aware representations.

Meanwhile, the loss function for face recognition has also been gradually improved \cite{schroff2015facenet, wen2016discriminative, zheng2018ring, liu2017sphereface, wang2018cosface, deng2019arcface, zhao2019regularface, duan2019uniformface}. Schroff {\em et al.} proposed triplet loss to minimize the distance between an anchor and a positive sample of the same identity, and maximize the distance between the same anchor and a negative sample with different identity \cite{schroff2015facenet}. Wen {\em et al.} presented center loss to learn a center for each identity and penalizes the distances between the features and their corresponding identity centers \cite{wen2016discriminative}. Zheng {\em et al.} applied ring loss to constrain a scaled unit circle, while maintaining convexity \cite{zheng2018ring}.

Although these previous methods have reached significant performance in face recognition, the deep networks still performs in a back-box manner, as we don't know what information the networks have learned.

\subsection{Expainability \& Interpretability}
The explainability and the discrimination power are two crucial properties of a deep network \cite{bau2017network, zhang2018interpretable}. At present, there is a trade-off between high explainability and strong discrimination power. Trough deep neural networks, high discrimination ability can be achieved but lack of the explainability \cite{zhang2018visual}. \\

% In this study, we sacrifice a few discrimination ability to obtained higher explainability. Nevertheless, we also reach a comparable result on our experiments.\\

\noindent{\bf Post-hoc Explanation via Visualization.  }
Post-hoc explanation aims to provide an understanding of what knowledge has been learned by networks, in an intuitive manner of humans \cite{du2019techniques}. Visualization is the most direct and intuitive way to explore the pattern hidden inside the deep network. \cite{zeiler2014visualizing} and \cite{simonyan2014deep} are two contemporary works, which proposed deconvolutional methods for visualizing knowledge learned by networks. \cite{simonyan2014deep} presented an gradient-based image-specific class saliency visualization method, which can visualize the spatial support of a particular class for a given input image. \cite{mahendran2015understanding, dosovitskiy2016inverting} shows a visualization method by inverting the image representations, which can not only be used for deep neural networks, but also for deep neural networks, but also for various features, such as histogram of oriented gradients (HOG) \cite{felzenszwalb2009object, dalal2005histograms}, local binary patterns (LBP) \cite{ojala2002multiresolution, ojala2000gray} and scale invariant feature transform (SIFT) \cite{lowe2004distinctive}.

In addition, visualization methods based on Class Activation Maps (CAMs) are also emerging. CAM based method was first introduced in \cite{zhou2016learning} to denote the discriminative regions of a particular category for a given input image. Subsequently, more and more improved CAMs, such as GradCAM \cite{selvaraju2017grad}, GradCAM++ \cite{chattopadhay2018grad}, ScoreCAM \cite{wang2020score} and so forth \cite{ramaswamy2020ablation, muhammad2020eigen}, are proposed.

Specifically, in face recognition, \cite {zhong2018deep} shows a method to find what difference between two similar-looking faces by a pair-wise patch-by-patch occlusion method, but it does not indicate what knowledge has been acquired by networks. \cite{zee2019enhancing} applied a CAMs method to generate heatmaps to indicate the active regions of different faces from different people; the results showed that the active regions tend to cover a whole face, so it does not provide useful information.

These post-hoc visualization methods have been successfully used in various fields of computer vision, like object classification \cite{liu2021group}, attacking \cite{zhang2021face}, biometrics \cite{qi2022research} and medical image processing \cite{qi2021parameter}, but they are difficult to generalize to face recognition. Therefore, model-intrinsic explanation methods are usually considered in face recognition. \\

\noindent{\bf Model-intrinsic Explanation via Visualization.  }
Model-intrinsic explanations are often derived from intrinsic, transparent, or white-box models \cite{linardatos2021explainable}, which has simple structures, such as decision trees or linear models. In this paper, we mainly focus on the methods to improve the explainability of network. These models have better partial explainability, to some extent, than traditional models. \cite{yin2019towards} proposed a spatial diversity loss and a feature diversity loss to enhance explainable representations, which reach an outstanding performance on three face recognition benchmarks, but it requires occluded faces before training. \cite{lin2021xcos} proposed a novel similarity metric, named explainable cosine (xCos), that can be plugged into various traditional networks to provide meaningful explanations on face recognition. Although it obtained comparable results, it did not reveal the knowledge learned inside the network. \\

In this study, we aim to constrain the deep network to learn an explainable facial part-based structural representation, without requirement for any manual annotation or occluded faces.

%\subsection{Part-based Structural Representation}
%It accords with human intuition that recognizes a face based on its part-based features. Many part-based face recognition methods have been proposed for decades \cite{berg2013poof, li2017probabilistic, li2001learning, sun2014deep, agarwal2004learning, yan2014face}. In deep learning, deep network can be regarded as an encoder to extract part-based feature from input face image.

\section{Method}
In this study, our goal is to guide the network to learn a facial part-based structural representation. Formally, we need to match the activation peaks of feature maps with the face part of the input face. In object detection and classification, the object always occupies adjacent pixels in the picture, but in face recognition, the face part, such as two eyes may be separated. To go a step further, we proposed a loss function to guide each channel of the feature map to contain maximum two activation peaks, and make an individual channel learn the same semantic parts.

As shown in Figure \ref{fig:channel_list}-(a), our activation templates contain all possible variations of activation patterns in the feature maps, which have zero, one, or two activation peaks. As we know, the representation of initial deep learning networks without training is random. Even for a well trained network, its representation is usually mixed with high-dimensional abstract features that humans cannot understand.

In this study, the task of the proposed ECLoss is to match the feature maps of the target layer within the network with a subset of activation templates $\mathbb{T}$. Therefore, we extend the definitions of mutual information in information theory from variable to sets of variables, which is a measure of the mutual dependence between the two sets. Our proposed ECLoss enables the network to learn the representation of activation templates by increasing the mutual information between activation templates and feature maps of the network. As shown in Figure \ref{fig:channel_list}-(c), our proposed ECLoss is applied to the last convolutional layer of the network to increase the mutual information $\text{MI}(\mathbb{T}; \mathbb{X})$ between activation templates $\mathbb{T}$ and feature maps $\mathbb{X}$. As a result, after training, the patterns of the feature maps $\mathbb{X}$ will match the activation templates $\mathbb{T}$

\begin{figure}[t]
	\includegraphics[width=\linewidth]{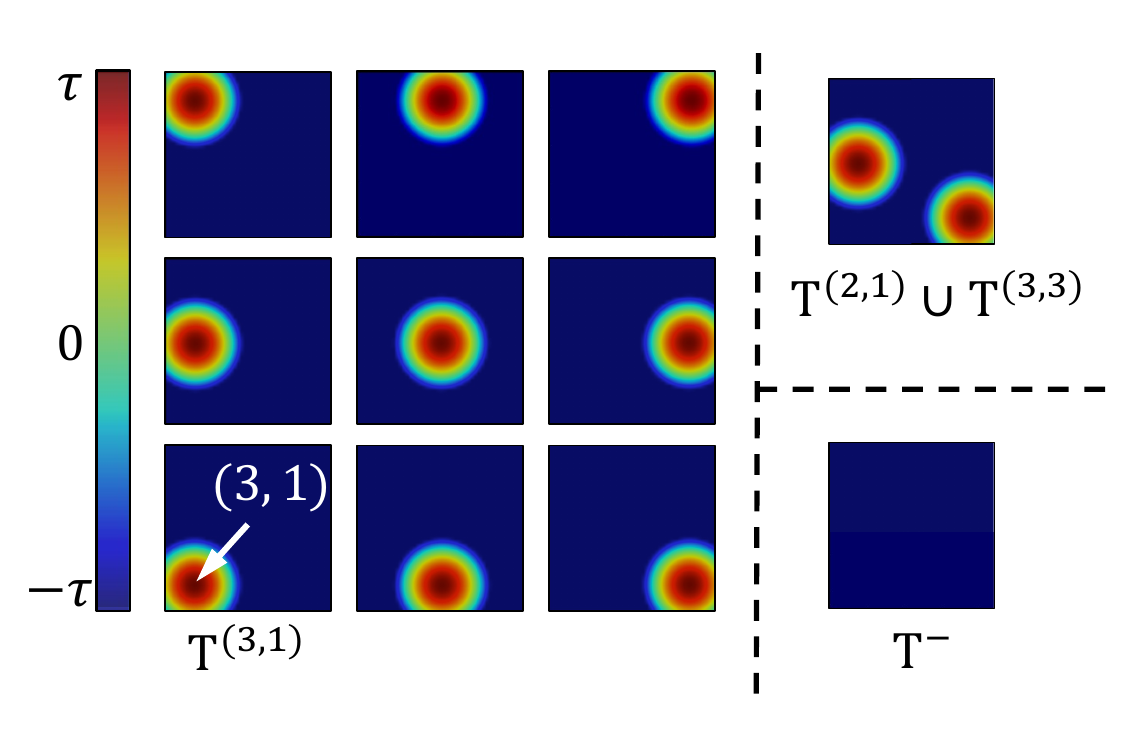}
	\centering
	\caption{{\bf(Left)} Examples of $3\times 3$ activation templates for ECLoss. {\bf(Top-right)} An example template with two activation peaks. {\bf(Bottom-right)} The negative template without any activation peak.}
	\label{fig:template_heatmap}
\end{figure}

The proposed ECLoss is a generic and model-agnostic method that can be used for various face recognition networks. For example, Figure \ref{fig:channel_list}-(c) shows that our generic channel loss function works on the last layer of the encoder, and can be used with other loss functions. Formally, in this work, our total loss function is defined as:
\begin{equation}
	\label{eq:total_loss}
	\begin{split}
		\text{Total Loss} = \alpha \text{Loss}(y, \hat{y}) + \beta \text{ECLoss}
	\end{split}
\end{equation}
where $\text{Loss}(y, \hat{y})$ indicates the loss between true subject ID label and the prediction, the $\alpha$ and $\beta$ are weighting coefficients.

\begin{figure}[t]
	\includegraphics[width=\linewidth]{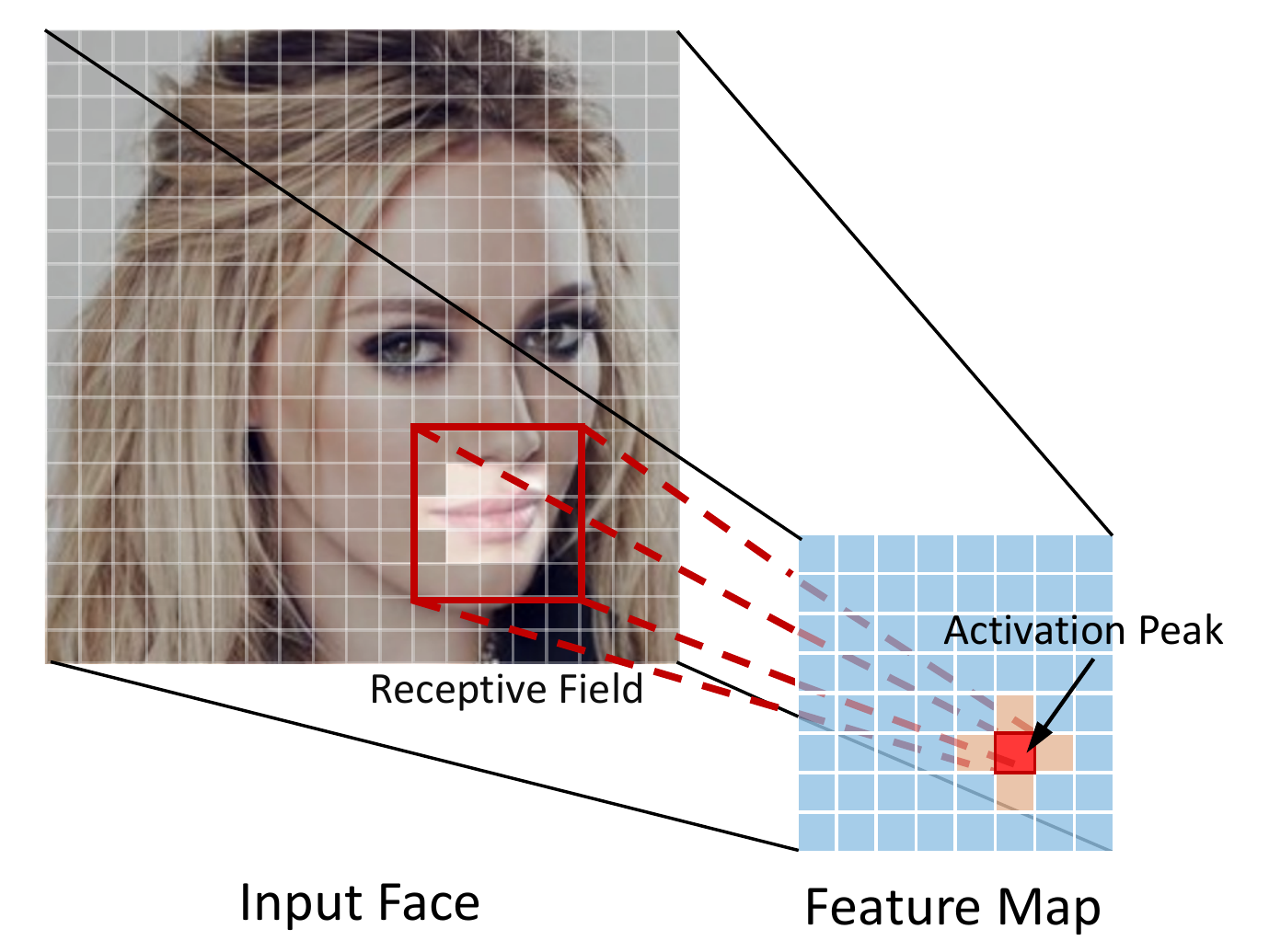}
	\centering
	\caption{Assuming that the face part of interest is the mouth, the pixels of the feature map corresponding to the mouth in its receptive field will be activated.}
	\label{fig:peak_example}
\end{figure}

\subsection{Activation Templates}
Given the face images $I$ with the size $s\times s$ as the input, the face recognition network with parameter $\theta$ can represent $I$ as the feature maps $\mathbb{X}$ from the target convolutional layer. In particular, $\mathbb{X}=\left \{x_1,\ x_2,\ \cdots ,\ x_i, \cdots,\ x_c\right \}$, where $x_i$ refers to the $i$-th channel of $\mathbb{X}$. \\

\noindent{\bf Activation templates with one peak. }
For an unknown face, the face parts may appear in any pixel of the input face image, {\em i. e.}, any pixel of the feature map may be activated if there is a face part of interest in the receptive field, as shown in Figure \ref{fig:peak_example}. 

Let $\text{T}^{(i, j)}$ denotes the activation template containing an activation peak in coordinate $(i,j)$, and $\text{T}^{(i, j)}_{(u, v)}$ indicates the value in coordinate $(u, v)$ of the activation template $\text{T}^{(i, j)}$. For each activation template, the value of the activation peak is $\tau$ and the value of other pixels extend outward the peak is defined as:
%Formally, the value $t$ of coordinate within the template $\text{T}_{\Delta_{(i, j)}}$ is defined as:
\begin{equation}
	\label{eq:tt}
	\begin{split}
		\text{T}^{(i, j)}_{(u, v)} = \tau\cdot\max(1-\frac{\left \|(u, v) - (i, j)\right \|_2}{r}, -1)
	\end{split}
\end{equation}
where the $\tau$ is a constant greater than $0$, $\left \| \cdot \right \|_2$ denote $L2$-Norm distance, $r$ is a constant that denotes the radius of the area greater than $0$. To go a step further, the one-peak activation template set $\mathbb{T}_1$ is defined as $\mathbb{T}_1 = \left \{ \  \text{T}^{(i, j)} \ | \ 1 \leq i, j\leq s \ \right \}$.

In addition, a negative template $\text{T}^-$ is defined to represent a feature map without any activation peaks, $\text{T}^-_{{(i, j)}} = -\tau$ for every coordinate $(i, j)$ within the template $\text{T}^-$. \\

\noindent{\bf Activation templates with two peaks. }
In order to match activation peaks with face parts, we allow a feature map to contain at most two activation peaks. In this situation, we define the two-peak templates to be the combination of any two one-peak templates, as shown in Figure \ref{fig:channel_list}. The combination $\text{T}^{(i_1, j_1)} \ \cup \ \text{T}^{(i_2, j_2)}$ of two one-peak templates $\text{T}^{(i_1, j_1)}$ and $\text{T}^{(i_2, j_2)}$ is defined as:
\begin{equation}
	\label{eq:mu}
	\begin{split}
	{\text{T}_{(u,v)}}=\max({\text{T}^{(i_1, j_1)}_{(u, v)}}, \ \text{T}^{(i_2, j_2)}_{(u, v)})
	\end{split}
\end{equation}
where ${\text{T}_{(u,v)}}$ is a two-peak template combining two one-peak template ${\text{T}^{(i_1, j_1)}_{(u, v)}}$ and $\text{T}^{(i_2, j_2)}_{(u, v)}$. Then, the two-peak template set $\mathbb{T}_2$ is defined as:
\begin{equation}
	\label{eq:mu}
	\begin{split}
		\mathbb{T}_2 = \left \{ \ \text{T}^{(i_1, j_1)} \ \cup \ \text{T}^{(i_2, j_2)}\ | \ 1 \leq i_1, i_2, j_1, j_2 \leq s \ \right \}
	\end{split}
\end{equation}

As shown in Figure \ref{fig:channel_list}, a set of activation templates $\mathbb{T}$ used for ECLoss contain all of the one-peak templates, two-peak templates and a negative template $\text{T}^-$, $\mathbb{T} = \mathbb{T}_1 \ \cup \  \mathbb{T}_2 \ \cup \ \left \{\text{T}^-\right \}$.

\subsection{Explainable Channel Loss}
In this study, our major goal is to make each dimension of the representation tend to represent a face part. Generally, before  training, the feature maps is a random output when the input face images go through the network, so the feature maps $\mathbb{X}$ and the locations of all face parts are unrelated. In this case, after network training, each dimension of the feature map belonging to $\mathbb{X}$ will be able to match activation templates $\mathbb{T}$. \\

\noindent{\bf Problem Definition: } Given a template set of activation pattern $\mathbb{T}$, and a feature map $\mathbb{X}$ of the target convolutional layer from the face recognition network, the task of ECLoss is to find parameters $\theta$ such that feature map $\mathbb{X}$ maximally represent a subset of the face part set $\mathbb{X}$.\\

By extending the definitions of mutual information in information theory from variable to sets of variables \cite{brillouin2013science, kullback1997information}, the mutual information between $\mathbb{X}$ and $\mathbb{T}$ is defined as: 
\begin{equation}
	\label{eq:MI}
	\begin{split}
		\text{MI} (\mathbb{T}; \mathbb{X}) & = \sum_{\text{T}\in \mathbb{T}} \sum_{x\in \mathbb{X}}\text{P}(\text{T}, x)\log({\text{P}(\text{T}, x) \over \text{P}(\text{T})\text{P}(x)}) \\
		& = \sum_{\text{T}\in \mathbb{T}} \sum_{x\in \mathbb{X}}\text{P}(x)\text{P}(\text{T}|x)\log({\text{P}(\text{T}|x) \over \text{P}(\text{T})})  
	\end{split}
\end{equation}
where $\text{\text{T}}(\mu)$ represents the prior probability of a activation template from $\mathbb{T}$. $\text{P}(x)$ denotes the probability of appearance of feature maps of different patterns, $\text{P}(x) = \sum_{\text{T} \in \mathbb{T}}\text{P}(\text{T})\text{P}(x|\text{T})$.  To simplify the problem, we assume that all templates are uniformly distributed, so the  $\text{P}(\text{T})$ is given as a constant of ${1\over |\mathbb{T}|}$. $\text{P}(\text{T}|x)$ is the conditional likelihood denoting the fitness between  $\text{T}$ and $x$:
\begin{equation}
	\label{eq:Px1}
	\begin{split}
		\text{P}(\text{T}|x)={{e^{tr(x \cdot \text{T})}}\over{\sum_{\hat{\text{T}} \in \mathbb{T}} e^{tr(x \cdot \hat{\text{T}})} }}
	\end{split}
\end{equation}
where $\cdot$ indicates hadamard product (element-wise product), $tr(*)$ indicates the trace of a matrix, so $tr(x \cdot \text{T})$ can be rewritten as $tr(x \cdot \text{T}) = \sum_i \sum_j {x_{ij} \text{T}_{ij}}$. 

To increase the mutual information between $\mathbb{T}$ and $\mathbb{X}$, ECLoss is defined as $\text{ECLoss}=-\text{MI}(\mathbb{T}; \mathbb{X})$. Then, the total loss function is defined as:
\begin{equation}
	\label{eq:total_loss_new}
	\begin{split}
		\text{Total Loss} = \alpha \text{Loss}(y, \hat{y}) - \beta \text{MI}(\mathbb{T}; \mathbb{X})
	\end{split}
\end{equation}

\subsection{Visualization of Structural Representation} \label{sec_vis}

Given the face images $I$ with the size $s\times s$ as the input, the face recognition network with parameter $\theta$ can represent $I$ as the feature maps $\mathbb{X}$ from the target convolutional layer after the ReLU operation. In particular, $\mathbb{X}=\left \{x_1,\ x_2,\ \cdots ,\ x_i, \cdots,\ x_c\right \}$, where $x_i$ refers to the $i$-th channel of $\mathbb{X}$. The feature map $x$ is a $n\times n$ matrix encoded by the face recognition network. In order to ensure the same resolution of $x$ as the input face, feature map $x$ will be resized to $s\times s$ by bi-linear upsampling.

To focus on the area of maximum activation, the area of top 10\% intensity will be highlighted, while other area will be discarded. A threshold $t$ is found to highlight the area of top 10\% intensity in the feature map $x$:
\begin{equation}
	\label{eq:hatx}
	\hat x_{(i,j)}=
	\begin{dcases}
			x_{(i,j)}, & x_{(i,j)} \geq t \\
			0, & \text{otherwise}
	\end{dcases}
\end{equation}

\begin{figure*}[p]
	\includegraphics[width=\linewidth]{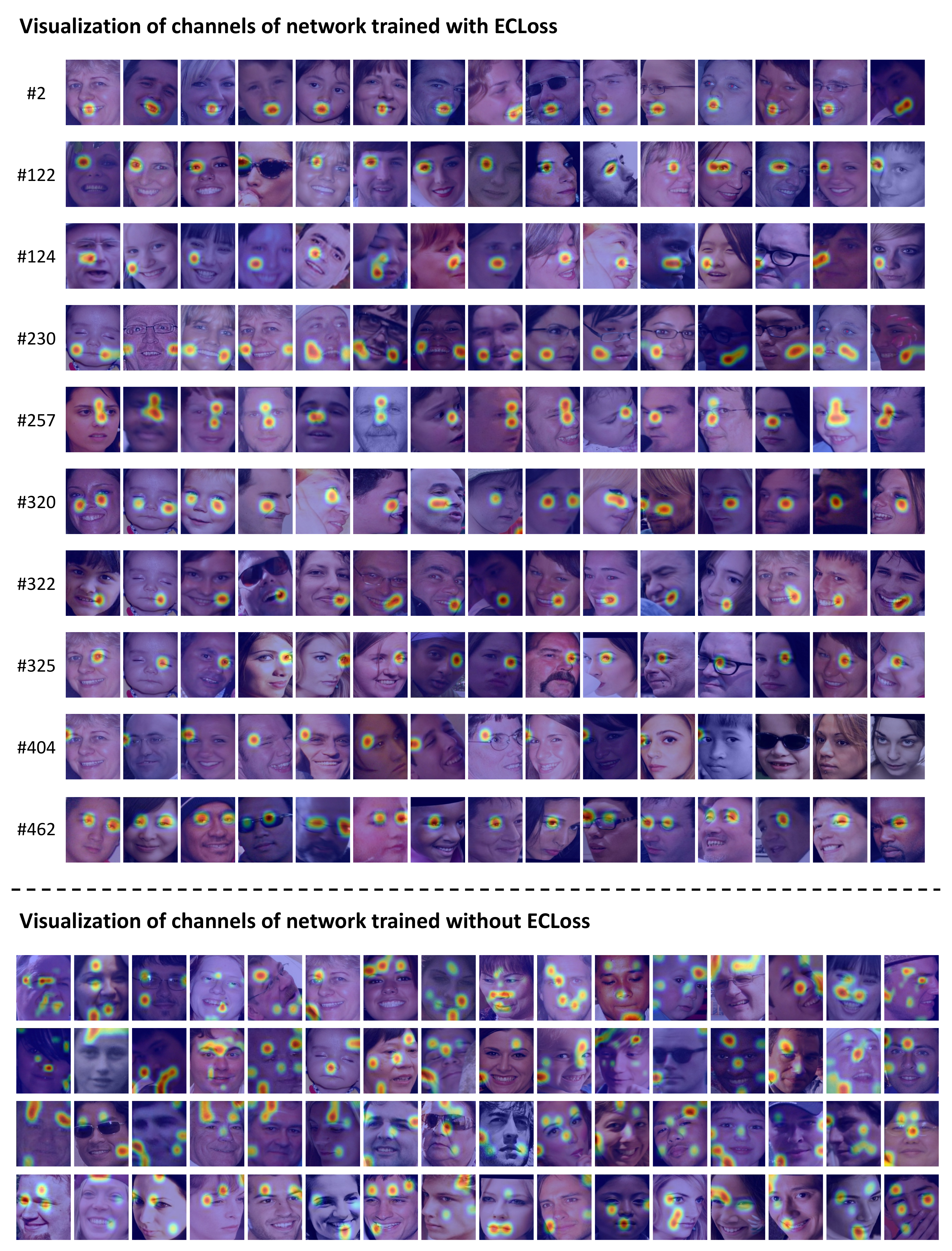}
	\centering
	\caption{Visualization for top $10\%$ activation. {\bf(Top)} The activation peaks of an individual channel tend to be located in the same face part, and each channel of feature maps contains at most two activation peaks. {\bf(Bottom)} Almost all feature maps that contain more than two activation peaks, the activation peaks have no specific meaning for face recognition.}
	\label{fig:fig_list}
\end{figure*}

Then the threshold $t$ is defined as:
\begin{equation}
	\label{eq:t}
	\begin{split}
		\hat t \equiv \arg\max_{t}\Big\{\text{P}(\hat x_{(i,j)} > t) \ge 0.1\Big\}
	\end{split}
\end{equation}
where $\text{P}$ is the probability when the condition is true. Some examples of visualization for top $10\%$ activation are shown on Figure \ref{fig:fig_list}, which are the result of transforming $\hat x$ into the heatmap style and superimposing it on the original face image.

\begin{figure*}[!thb]
	\includegraphics[width=\linewidth]{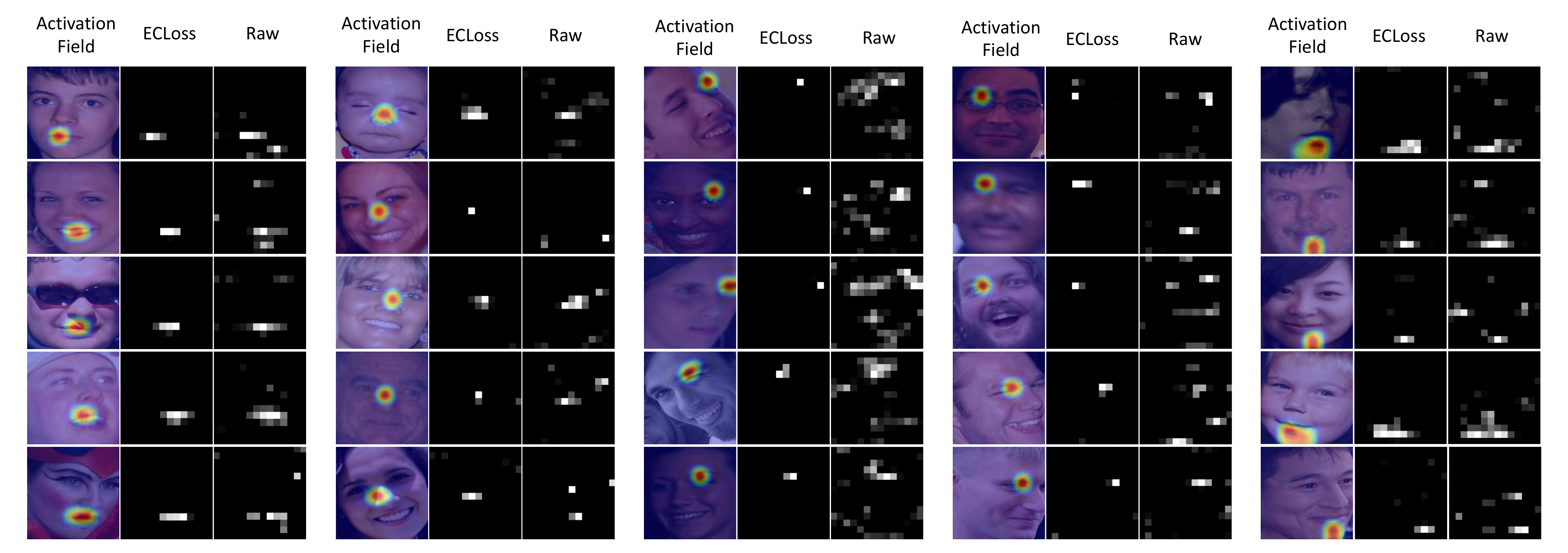}
	\centering
	\caption{Comparison of the feature map between networks with and without ECLoss. For the same input face, the feature map of the explainable network trained with ECLoss has at most two activation peaks, and usually activates a certain face part. Instead, the feature map of the original network has more noises, and the activation peaks are located randomly on different parts.}
	\label{fig:cmp_raw}
\end{figure*}

\section{Experiment}
\subsection{Dataset \& Pre-processing}
\noindent{\bf  Training Dataset }  The Glint360K \cite{an2021partial} is a large scale dataset merged from Celeb-500K \cite{cao2018celeb} and MS-Celeb-1M \cite{guo2016ms}, containing 17 million face images from 360K individuals. All face images of the Glint360K are downloaded from Internet and then pre-processed to size of $112 \times 112$ by 5 facial marks predicted by the face localisation method RetinaFace \cite{deng2020retinaface}.

\noindent{\bf  Testing Dataset } 
For testing robustness of face recognition, we explore various face verification datasets: LFW \cite{huang2008labeled}, CFP-FF \cite{sengupta2016frontal} and CALFW \cite{zheng2017cross}. Besides, in this study, many explainability metrics are assessed using several datasets, such as VGGFace\cite{parkhi2015deep}, AgeDB \cite{moschoglou2017agedb}, Helen \cite{le2012interactive}, 300-W\cite{sagonas2016300}, LaPa \cite{liu2020new}, AFW \cite{zhu2012face} and LFPW \cite{belhumeur2013localizing, sagonas2013semi}. All datasets used in training and testing are listed in Table \ref{tab:datasets}.\\

\begin{table}[htb]
	\centering

	\renewcommand\arraystretch{1.5}
	\setlength{\tabcolsep}{3pt}
	\caption{\label{tab:datasets} The list of face datasets used for training and testing}
	\vspace{0.8em}
	\begin{tabular}{p{28pt}<{\centering}p{54pt}<{\centering}p{26pt}<{\centering}p{50pt}<{\centering}p{58pt}<{\centering}}
		\toprule[2pt]
		\textbf{Stage}& 
		\textbf{Dataset}& 
		\textbf{Landmark}& 
		\textbf{Pose}&
		\textbf{Image / Pairs}\\
		\midrule[1.2pt]
		\multirow{1}*{Training}
		&\small{Glint360K \cite{an2021partial}}&-&Near-frontal&17,091,657 / -\\
		\midrule[1.2pt]
		\multirow{9}*{Testing}
		&LFW \cite{huang2008labeled}&-&Near-frontal&- / 6,000\\
		\cline{2-5}
		&CFP-FF \cite{sengupta2016frontal}&-&Near-frontal&- / 7,000\\
		\cline{2-5}
		&CALFW \cite{zheng2017cross}&-&Near-frontal&- / 3,000\\
		\cline{2-5}
		&\small{VGGFace \cite{parkhi2015deep}}&-&Near-frontal&982,803 / -\\
		\cline{2-5}
		&AgeDB \cite{moschoglou2017agedb}&-&Near-frontal&16, 488 / -\\
		\cline{2-5}
		&300-W \cite{sagonas2016300}&68&Near-frontal&3,148 / -\\
		\cline{2-5}
		&AFW \cite{zhu2012face}&68&$[-45^\circ,\ 45^\circ]$&750,000 / -\\
		\cline{2-5}
		&Helen \cite{le2012interactive}&68&$[-90^\circ,\ 90^\circ]$&2,000 / -\\
		\cline{2-5}
		&LaPa \cite{liu2020new}&106&$[-90^\circ,\ 90^\circ]$&22,000 / -\\
		\cline{2-5}
		&LFPW \cite{belhumeur2013localizing}&68&$[-90^\circ,\ 90^\circ]$&1,432 / -\\
		\bottomrule[2pt]
	\end{tabular}
	
\end{table}

\noindent{\bf  Pre-processing } 
Different faces have almost the same characteristics, such as the position, size and shape of facial components. For face recognition, face alignment is an essential prerequisite, which can align facial components, e.g., eye, nose, mouth, and contour to a standard facial space. In this study, in order to reduce the influence of the structural representation by the standard facial space, the data used for training and testing are not aligned. All face images used for training and testing are detected and cropped by MTCNN \cite{xiang2017joint}, and resized to $224 \times 224$. \\

\subsection{Implementation Details } 
We trained three explainable networks based on VGG13 \cite{simonyan2014very}, VGG16 \cite{simonyan2014very}, and VGG19 \cite{simonyan2014very} for face recognition in the Glint-360K dataset. And original VGG13, VGG16, and VGG19 were trained on the same dataset and settings for comparison. As the skip connections of the residual networks mix different pattern of multiple layers in a single feature map \cite{he2016deep, zhang2020interpretable}, we do not use residual networks and skip connection in this study. In these experiments, the input face images are detected by MTCNN and resized to $224 \times 224$, without any face alignment. 

For activation templates, we found in our experiments that the results are not sensitive to the value of $\tau$. As ECloss might overflow when the value of $\tau$ is big, we set $\tau$ to a small number, {\em i.e.} $t = 0.001$, and $r = 4$. For total loss function of $\text{Total Loss} = \alpha \text{Loss}(y, \hat{y})  - \beta \text{MI}(\mathbb{T}; \mathbb{X})$, we set the $\alpha$ as a constant of $1$, and $\beta$ is set as $1e^{-5}$ initially and then automatically changed by the program to reduce the total loss.

In the VGG architecture, the shape of last convolutional layer is $14 \times 14$, {\em i. e.}, for each channel, tens of thousands of activation templates can be generated (nearly $14^4$ two-peak templates, $14^2$ one-peak templates and $1$ negative template). To reduce the complexity of calculation, we take $400$ templates evenly from original template set $\mathbb{T}^{ori}$ as a new set $\mathbb{T}$ used for all experiments.

For mutual information $\text{MI}(\mathbb{T}; \mathbb{X})$ between $\mathbb{T}$ and $\mathbb{X}$, we assume that all templates of $\mathbb{T}$ are uniformly distributed, so we set $|\mathbb{T}| = 400$, $\text{P}(\text{T}) = {1\over |\mathbb{T}|} = 0.0025$.  Because $\text{P}(x) = \sum_{\text{T}}\text{P}(\text{T})\text{P}(x|\text{T}), \  \text{T} \in \mathbb{T}$ is computed using numerous prior feature map requiring large amounts of memory. we approximate $\text{P}(x)$ as $ \sum_{\text{T}}\text{P}(\text{T})\mathbf{E}_x\left[{{e^{tr(x \cdot \text{T})}}\over{\sum_{\hat{\text{T}}} e^{tr(x \cdot \hat{\text{T}})} }}  \right]$, where $\mathbf{E}[\cdot]$ indicates the mathematic expectation, $tr(\cdot)$ indicates the trace of a matrix, and $\text{P}(\text{T}) = 0.0025$.

All experiments are conducted on a workstation with eight NVIDIA Tesla V100 GPUs. Moreover, the learning rate is set to 0.03 and the batch size is set to 64.

\section{Result \& Analysis}

\subsection{Visualization of Activation} 
As mentioned in the section \ref{sec_vis}, we visualized the receptive fields of activation of a channel.  Previous studies in \cite{bau2017network, zhou2018interpreting} and \cite{zhang2018interpretable} have introduced this method to compute the receptive fields on a given feature map.

For both networks trained with and without ECLoss, we visualize the top $10\%$ activation values of the last convolutional layer. As shown in Figure \ref{fig:fig_list}, the maximal activation area in the network trained with ECLoss can focus on a certain facial part, like face, cheek, nose, {\em etc.}, but the original network randomly focus on some unexplainable semantic information. Moreover, for explainable networks trained with ECLoss, these clear disentanglements of structural part-based representation can easily help people to know what knowledge the network has learned, which is helpful for quantifying the contribution of different face parts to the network.

We found that, within the explainable face recognition networks trained with ECLoss, an individual layer usually detected a certain face part even if the input face images are from different people. For example, as shown in Figure \ref{fig:fig_list}, the $2$nd channel detected the mouth, the $122$nd channel detected the left eye, the $257$th channel detected the nose, the $325$th channel detected the right eye, the $462$nd channel detected the eyes, {\em etc.} In addition, it not only detected face components, but also learned some patterns that make sense to the network. For example, the $124$th channel learned the area to the right of the nose, and the $404$th channel learned the area to the right of the right eye.

In addition, we also compare the feature map of the explainable network and the original network in Figure \ref{fig:cmp_raw}. Given the same face as input, our explainable networks usually have at most two activation peaks in a feature map, and the activation peaks tend to activate a certain face part. Instead, the activation peaks of the feature map of the original network usually contain a lot of noises, and are randomly located on different areas.

\begin{table*}[htb]
	\centering
	\renewcommand\arraystretch{1.5}
	\setlength{\tabcolsep}{3pt}
	\caption{\label{tab:Part} The evaluation metric of part explainability on various dataset: LaPa, AFW, Helen, LFPW and 300-W.}
	\vspace{0.8em}
	\begin{tabular}{p{100pt}p{65pt}<{\centering}p{65pt}<{\centering}p{65pt}<{\centering}p{65pt}<{\centering}p{65pt}<{\centering}}
		\toprule[2pt]
		\textbf{Network}& 
		\textbf{LaPa}& 
		\textbf{AFW}&
		\textbf{Helen}&
		\textbf{LFPW}&
		\textbf{300-W}\\
		\bottomrule[1.2pt]
		VGG13\small{ + Softmax} & 0.1642 & 0.1612 & 0.1646 & 0.1625 & 0.1554\\
		\hline
		VGG13 + ECLoss & {\bf 0.1710} & {\bf 0.1687} & {\bf 0.1711} & {\bf 0.1702} & {\bf 0.1623}\\
		\midrule[1.2pt]
		VGG16\small{ + Softmax} & 0.1479 & 0.1447 & 0.1530 & {\bf 0.1443} & {0.1396}
		\\
		\hline
		VGG16 + ECLoss  & {\bf 0.1632} & {\bf 0.1490} & {\bf 0.1566} & { 0.1431} & {\bf 0.1443}\\
		\bottomrule[1.2pt]
		VGG19\small{ + Softmax} & 0.0424 & 0.0396 & 0.0396 & 0.0376 & 0.0381\\
		\hline
		VGG19 + ECLoss  & {\bf 0.0432}  & {\bf 0.0426} & {\bf 0.0466} & {\bf 0.0430} & {\bf 0.0376}\\
		\bottomrule[2pt]
	\end{tabular}
	
\end{table*}

  \begin{figure}[t]
 	\includegraphics[width=\linewidth]{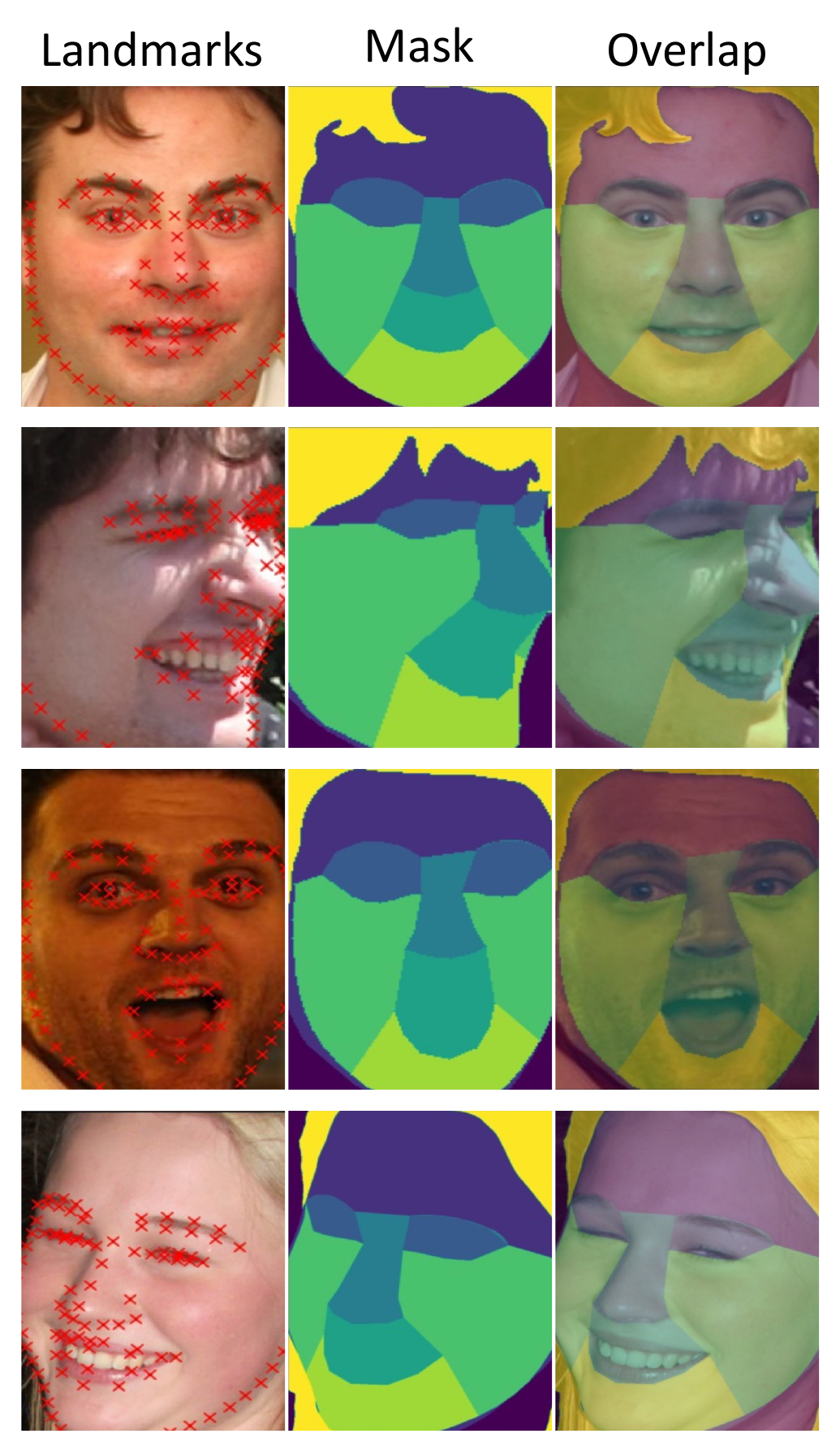}
 	\centering
 	\caption{Re-annotation by landmarks for evaluation of part explainability. {\bf (Left Column)} Original landmarks on face from face parsing dataset. {\bf (Middle Column)} Re-annotation mask by landmarks containing hair, forehead, cheek, eyes, nose, mouth and jaw. {\bf (Right Column)} Visualization of mask overlapped the face.}
 	\label{fig:mask}
 \end{figure}
 
\subsection{Quantitative Analysis}
\subsubsection{Face Verification}

Face verification mainly computes the similarity between a reference and a probe face image to decide if these face images are of the same identity \cite{crosswhite2018template}. In the evaluation of face verification, LFW \cite{huang2008labeled}, CFP-FF \cite{sengupta2016frontal} and CALFW \cite{zheng2017cross} datasets are used to assess performance of the explainable face recognition networks and the original networks.

In all experiments, we did not perform face alignment. As shown in Table 2, compared to the original networks, the explainable network with the proposed ECLoss attained a significant improvement. The VGG13 network trained with ECLoss achieved 0.2\%, 0.13\% and 0.58\% higher accuracy on LFW, CFP-FF, and CALFW datasets, respectively. The results of VGG16 with ECLoss are 0.88\%, 2.08\% and 2.6\% higher than the original network. The accuracy of VGG19 with ECLoss is 0.79\%, 0.26\% and 5.01\% higher than the original network.

\begin{table}[htb]
	\centering
	\renewcommand\arraystretch{1.5}
	\setlength{\tabcolsep}{3pt}
	
	\caption{\label{Face} Face verification accuracy on LFW, CFP-FF and CALFW datasets without face alignment. Compared to the original networks, the explainable network with the proposed ECLoss achieves a significant improvement.}
	\vspace{0.8em}
	\begin{tabular}{p{80pt}p{45pt}<{\centering}p{45pt}<{\centering}p{45pt}<{\centering}}
		\toprule[2pt]
		\textbf{Network}& 
		\textbf{LFW}& 
		\textbf{CFP-FF}&
		\textbf{CALFW}\\
		\bottomrule[1.2pt]
		VGG13\small{ + Softmax} & 91.95\% & {79.57\%} & 76.55\% \\
		\hline
		VGG13 + ECLoss & \textbf{92.15\%} & \textbf{79.70\%} & \textbf{77.13\%} \\
		\midrule[1.2pt]
		VGG16\small{ + Softmax}& 93.20\% & 78.59\% & 78.10\% \\
		\hline
		VGG16 + ECLoss & \textbf{94.08\%} & \textbf{80.67\%} & \textbf{80.70\%} \\
		\bottomrule[1.2pt]
		VGG19\small{ + Softmax} & 90.96\% & 75.90\% & 70.97\% \\
		\hline
		VGG19 + ECLoss & \textbf{91.75\%} & \textbf{76.16\%} & \textbf{75.98\%} \\
		\bottomrule[2pt]
	\end{tabular}
	
\end{table}

\begin{table*}[htb]
	\centering
	\renewcommand\arraystretch{1.5}
	\setlength{\tabcolsep}{3pt}
	
	\caption{\label{tab:loca} The evaluation metric of location consistency on various dataset: LaPa, AFW, Helen, LFPW and 300-W.}
	\vspace{0.8em}
	\begin{tabular}{p{100pt}p{65pt}<{\centering}p{65pt}<{\centering}p{65pt}<{\centering}p{65pt}<{\centering}p{65pt}<{\centering}}
		\toprule[2pt]
		\textbf{Network}& 
		\textbf{LaPa}& 
		\textbf{AFW}&
		\textbf{Helen}&
		\textbf{LFPW}&
		\textbf{300-W}\\
		\bottomrule[1.2pt]
		VGG13\small{ + Softmax} & 0.02537 & 0.01971 & 0.02082 & 0.02187 & 0.02122\\
		\hline
		VGG13 + ECLoss  & {\bf 0.04029} & {\bf 0.03654} & {\bf 0.03661} & {\bf 0.03796} & {\bf 0.03770}\\
		\midrule[1.2pt]
		VGG16\small{ + Softmax}  & 0.02743 & 0.03516 & 0.03032 & {\bf 0.04230} & 0.02867\\
		\hline
		VGG16 + ECLoss  & {\bf 0.02898} & {\bf 0.03589} & {\bf 0.03327}& 0.04175 & {\bf 0.03417}\\
		\bottomrule[1.2pt]
		VGG19\small{ + Softmax} & 0.08075 & 0.08307 & 0.08767 & 0.08217 & 0.08450\\
		\hline
		VGG19 + ECLoss  & {\bf 0.09382} & {\bf 0.09503} & {\bf 0.09592} & {\bf 0.09453} & {\bf 0.09418}\\
		\bottomrule[2pt]
	\end{tabular}
	
\end{table*}

\begin{table*}[htb]
	\centering
	\renewcommand\arraystretch{1.5}
	\setlength{\tabcolsep}{3pt}
	
	\caption{\label{act}The evaluation metric of \textbf{activation robustness} (Means ± Standard Error) on various dataset: VGGFace, LFW, CFP, AgeDB, Helen and LaPa.}
	\vspace{0.8em}
	\begin{tabular}{p{80pt}p{65pt}<{\centering}p{65pt}<{\centering}p{65pt}<{\centering}p{65pt}<{\centering}p{65pt}<{\centering}p{65pt}<{\centering}}
		\toprule[2pt]
		\textbf{Network}& 
		\textbf{VGGFace}& 
		\textbf{LFW}&
		\textbf{CFP}&
		\textbf{AgeDB}&
		\textbf{Helen}&
		\textbf{LaPa}\\
		\bottomrule[1.2pt]
		VGG13\small{ + Softmax} & 3.733 ± 0.094 & 3.945 ± 0.022 & 3.580 ± 0.205 & 3.984 ± 0.147 & 4.070 ± 0.164 & 3.918 ± 0.236\\
		\hline
		VGG13 + ECLoss & \textbf{2.596 ± 0.050} &  \textbf{2.784 ± 0.077} &  \textbf{2.495 ± 0.126} &  \textbf{2.718 ± 0.221} &  \textbf{2.816 ± 0.105} &  \textbf{2.889 ± 0.166}\\
		\midrule[1.2pt]
		VGG16\small{ + Softmax} & 3.714 ± 0.081 &  3.994 ± 0.041 & 3.994 ± 0.041 & 3.889 ± 0.167 & 3.946 ± 0.144 & 3.847 ± 0.195\\
		\hline
		VGG16 + ECLoss  &  \textbf{3.144 ± 0.086} & \textbf{3.272 ± 0.013} & \textbf{3.210 ± 0.078} & \textbf{3.429 ± 0.165} & \textbf{3.242 ± 0.151} & \textbf{3.395 ± 0.186}\\
		\bottomrule[1.2pt]
		VGG19\small{ + Softmax} & 2.987 ± 0.111 & 2.965 ± 0.247 & 2.646 ± 0.069 & 3.053 ± 0.123 & 3.083 ± 0.148 & 3.068 ± 0.042\\
		\hline
		VGG19 + ECLoss  & \textbf{1.672 ± 0.108} & \textbf{1.743 ± 0.164} & \textbf{1.493 ± 0.037} & \textbf{1.775 ± 0.183} & \textbf{1.829 ± 0.053} & \textbf{1.849 ± 0.065}\\
		\bottomrule[2pt]
	\end{tabular}
	
\end{table*}

\subsubsection{Part Explainability}

This evaluation metric of part explainability was proposed by Bau {\em et al.} \cite{bau2017network} and Zhang {\em et al.} \cite{zhang2018interpretable}, which measures the object-part explainability of channels from a network.

For assessing part explainability of various network, five face parsing datasets of LaPa  \cite{liu2020new}, AFW \cite{zhu2012face}, Helen \cite{le2012interactive}, LFPW \cite{belhumeur2013localizing} and 300-W \cite{sagonas2016300} datasets are used. The details of these datasets are shown in Table \ref{tab:datasets}. Before evaluation of part explainability, as shown in Figure \ref{fig:mask}, we generate facial component masks for these face datasets by their landmarks, which contain hair, forehead, cheek, eyes, nose, mouth, and jaw. Formally, a facial components set is defined as $\mathbb{K} = \left \{\text{hair, forehead, cheek,}\ etc. \right \}$.

Given a face recognition network, let $I$ donate the face image input to the network, $x$ represents a channel of the feature maps $\mathbb{X}$ from the last convolutional layer of the encoder after the activation function of ReLU. The feature map $x$ is first resized to $s\times s$ by bi-linear upsampling, where $s$ is the size of the input image. A threshold $t$ is then applied to find out the top $10\%$ activation area using Equation \eqref{eq:t}. Then, the fitness of feature map $x$ regarding to the facial component $k \in \mathbb{K}$ is computed by 
\begin{equation}
	\label{eq:IOU}
	\begin{split}
		\text{IoU}_{\text{Mask}_k, x}= \frac{\text{P}\big[\text{Mask}_k \ \cap \ x\big]}{\text{P}\big[\text{Mask}_k \ \cup \ x\big]}
	\end{split}
\end{equation}
where $\text{P}$ is the probability when the condition is true, and $\text{Mask}_k$ is the binary mask that represents the ground-truth of the facial component $k$, which is derived from our re-annotation of various datasets shown on Figure \ref{fig:mask}.

For evaluation of part explainability, we compute $\text{PE}_{I} = \mathbf{E}_{x \in \mathbb{X}}\big[\max_k( \text{IoU}_{\text{Mask}_k, \ x})\big]$ for each input face image. As shown on Table \ref{tab:Part}, for each network, we conduct several independent experiments, where a random selection of 100  face images from each face parsing dataset is used to calculate the part explainability.

% 分析结果数据

\subsubsection{Location Consistency}
The metrics of location consistency assess the consistency of the location of the activation peaks \cite{zhang2018interpretable, bau2017network}, which indicate whether the channel of feature maps is able to detect the same face part regardless of the input image, as shown in Figure \ref{fig:fig_list}.

For evaluation of location consistency, we first find the face component $\hat k$ with maximum correspondence by $\hat k = \arg\max_k\big( \text{IoU}_{\text{Mask}_k, x}\big)$. Therefore, for the $i$-th channel of feature maps $x_i$, we compute its corresponding face part $\hat k$ for different input images $I$.

Given a channel's index $i$, $A_{i,k}$ indicates the probability of the feature map $x_i$ corresponding to the $k$-th face part, {\em i. e.}, for an arbitrary channel's index $i$, $\sum_{k\in\mathbb{K}}{{A}_{i,k}} = 1$. For the $i$-th channel, we use $S_i = \text{var}[A_{i,k}] = {\sqrt{\sum{(A_{i,k}-{1\over|\mathbb{K}|})}}\over |\mathbb{K}|}, k\in\mathbb{K}$ to assess its consistency. For a network architecture, we use $\text{LS} = \mathbf{E}\big[S_i\big] = {\sum_{i=1}^{c}S_i\over c}$ to evaluate location consistency of the network.

The results are shown in Table \ref{tab:loca}. For each network, we conduct several independent experiments, using a random selection of 100 face images from each face parsing dataset (LaPa  \cite{liu2020new}, AFW \cite{zhu2012face}, Helen \cite{le2012interactive}, LFPW \cite{belhumeur2013localizing} and 300-W \cite{sagonas2016300}).

\subsubsection{Activation Robustness}

The main goal of the proposed ECLoss is to force a feature map to contain at most two activation peaks. As shown in Figure \ref{fig:fig_list} and Figure \ref{fig:cmp_raw}, compared to the feature maps of explainable networks trained with ECLoss, the feature maps of original networks usually involve more than two activation peaks.  Therefore, when the number of activation peaks is smaller, the activation pattern is more robust.

In simple terms, the purpose of Activation Robustness is to evaluate the average number of peaks in the feature map. Given a feature map $x$ after ReLU operation from the last convolutional layer. Then, a contour method \cite{prime2013contour, prime2013practical} will be used to compute peak prominence on the feature map $x$, which uses enclosing contours to define the local maxima.

In the experiments of activation robustness, for VGG13,  VGG16,  VGG19,  explainable VGG13, explainable VGG16, and explainable VGG19, we conduct 5 independent experiments to calculate the mean and standard deviation of the number of activation peaks. The results of these experiments have been shown in Table \ref{act}, which demonstrate that our proposed ECLoss achieves better robustness on the six datasets of VGGFace \cite{parkhi2015deep}, LFW \cite{huang2008labeled}, CFP \cite{sengupta2016frontal}, AgeDB \cite{moschoglou2017agedb}, Helen \cite{le2012interactive} and LaPa \cite{liu2020new}.

% 分析结果数据

\section{Conclusion}
In this paper, we proposed an Explainable Channel Loss (ECLoss) to construct an explainable face recognition network by forcing the target convolution layer to learn the facial part-based representation. Each channel of the feature maps represents a certain face part, without any manual annotation or additional occlusion datasets. We assess the proposed ECLoss by three traditional deep learning architectures VGG13 \cite{simonyan2014very}, VGG16 \cite{simonyan2014very}, and VGG19 \cite{simonyan2014very} on dozens of datesets like Glint360K \cite{an2021partial}, LFW \cite{huang2008labeled}, CFP \cite{sengupta2016frontal}, CALFW \cite{zheng2017cross}, VGGFace \cite{parkhi2015deep}, AgeDB \cite{moschoglou2017agedb}, 300-W \cite{sagonas2016300}, AFW \cite{zhu2012face}, Helen \cite{le2012interactive}, LaPa \cite{liu2020new} and LFPW \cite{belhumeur2013localizing}. All of our experiments have shown that our proposed ECLoss can achieve significant explainability metrics while improving the performance of face recognition.

\section{Acknowledgements}
This research was supported by National Natural Science Foundation of China under grant no. 91959108.

{\small
	
	\bibliographystyle{ieee_fullname}
	\bibliography{references}
}

\end{document}